# A Survey on Importance of Homophones Spelling Correction Model for Khmer Authors

Seanghort BORN†‡      Madeth MAY†      Claudine PIAU-TOFFOLON†
Sébastien IKSAL†

†Laboratoire d'Informatique de l'Université du Mans (LIUM), Le Mans University, France
‡Cambodia Academy of Digital Technology (CADT), Cambodia
{seanghort.born.etu, madeth.may, claudine.piau-toffolon, sebastien.iksal}@univ-lemans.fr
seanghort.born@cadt.edu.kh

**Abstract**

Homophones present a significant challenge to authors in any languages due to their similarities of pronunciations but different meanings and spellings. This issue is particularly pronounced in the Khmer language, rich in homophones due to its complex structure and extensive character set. This research aims to address the difficulties faced by Khmer authors when using homophones in their writing and proposes potential solutions based on an extensive literature review and survey analysis. A survey of 108 Khmer native speakers, including students, employees, and professionals, revealed that many frequently encounter challenges with homophones in their writing, often struggling to choose the correct word based on context. The survey also highlighted the absence of effective tools to address homophone errors in Khmer, which complicates the writing process. Additionally, a review of existing studies on spelling correction in other languages, such as English, Azerbaijani, and Bangla, identified a lack of research focused specifically on homophones, particularly in the Khmer language. In summary, this research highlights the necessity for a specialized tool to address Khmer homophone errors. By bridging current gaps in research and available resources, such a tool would enhance the confidence and accuracy of Khmer authors in their writing, thereby contributing to the enrichment and preservation of the language. Continued efforts in this domain are essential for ensuring that Khmer can leverage advancements in technology and linguistics effectively.

**Keywords:** Khmer language, Homophones

## 1 Introduction

A language is not just for communication, but it represents an identity of a group of people or even a nation. Each language has its own specialties and difficulties. Spelling mistakes are a challenge to users in all languages. A study on how spelling errors impact on perceptions of authors from the Department of Psychology of Central Missouri State University in the United States, indicated that when a reader encountered spelling issues, they may assume that it represents to the overall ability of the authors [1]. The same study emphasized that spelling errors resulted in typing assurance, misunderstandings, and even social disruptions. There are different spelling issues such as vocabulary misspellings, grammatical errors, and misused-context terms. In [2], spelling errors are categorized into 2 main types: typographic errors or non-word errors, and cognitive errors or real-word errors. Typographic errors happen when the correct spelling of a term is known (example: become) yet it is mistyped (example: beco*em*). Non-word errors are classified into 4 groups – insertion (example: becom*ea)*, deletion (example: *bc*ome), substitution (example: be*cu*me), and transposition (example: beco*em*) [3], [4]. These errors may happen due to keyboard errors, confused key typed, or the author not sure about the correct words [4].

In contrast, real-word errors happen when the correct spelling is unknown. For example, "too" is an errors words of "two" in the sentence: "I met *too* (two) boys this afternoon." It normally happens when the correct word is unknown and replaced by another familiar term. Homophone is a cognitive error that authors often find as a challenge in their writing. It is one of the top 5 misspelling mistakes that the authors frequently encounter. Not different from others, Khmer authors also endure the same issue. Due to the complexity of the language, and less related research to Khmer homophone spelling correction, it remains a concern to authors in their writing.

### 1.1 Homophones

Homophones refer to two or more words having the same pronunciation, but different spelling or meaning. For instance, "there" and "their", they have the same pronunciation [ðɛər], but different spellings and meanings. The term "there" refers to in or at a place, while "their" is a form of possessive pronoun of "them" that means belong to them. Homophones are frequently confused by authors, and they cause text confusion to readers sometimes. Additionally, to identify homophones spelling error is another level of linguistic difficulty because each word in the sentence seems correctly spelled, yet there might be some mistakes happen when we carefully review based on the context that those words located.

### 1.2 Khmer Language

Khmer is the official language of Cambodians which is used by some twenty million people worldwide [5]. It has up to 114 characters which ranged from U+1780 to U+17FF in the Unicode table [6]. It contains 74 letters and noted as the longest alphabets in Guiness World Record book in 1995 [7], [8]. Khmer is one of the oldest written languages from Mon-Khmer family [9]. It is written from left to right, and top to bottom, with multiple levels of characters stacking possible and use diacritics that further enhance the pronunciation of words. Due to the richness of the language, there are many homophones in Khmer. However, it remains a few research concerns to these problems, and it becomes challenges to authors in their Khmer writing. Our research aims to define difficulties of authors in homophone utilization in Khmer writing and propose a solution for the concern based on the literature review of related works.

In the following sections, we will address the research problems in Section 2, conduct a literature review in Section 3, explore the methodology in Section 4, and present the results and discussion in Section 5, with the conclusion in Section 6.

## 2 Research Problems

Homophones are a critical challenge to authors. Due to their similarity, users easily make homophone mistakes in their writing. In Khmer there are plenty of homophones that cause confusion in authors' writing. For instance, two following terms: "ខ្លា" [kla] means "tiger", and "ក្លា" [kla] means "brave". They both have the same pronunciation [kla] yet completely different meanings which are often confused by users in their writings. There are not only these two words, but many other terms that are easily distracted, and cause homophone mistakes. Nevertheless, the assistant tool related to this concern that could enhance authors' works like rich-resource languages such as English and French does not exist yet. In addition, research works related to this matter is also still very few that cause difficulty in application development to assist authors' works.

In conclusion, homophones are still a challenge to authors in Khmer writings, yet there is no powerful existing tool to assist their work nor research study to this concern. It needs our efforts to be involved with this matter to make better use of homophones in Khmer and as a state of the art to other related works.

## 3 Literature Review

Spelling error is a crucial part in text communication. It does not only interrupt readers' reading, but it might also cause readers to make assumptions about the ability of the authors. A research from the Department of Psychology of Central Missouri State University in the United States specified that when readers discover spelling errors, they may believe that they reflect to the author's overall ability [1]. Spelling error is not just a tiny problem that can be overlooked. But it does affect our communication. Furthermore, to handle terms that have the same pronunciation, it is another difficulty level to authors that deserve our researchers' attention with this issue.

There are different studies that have been conducted related to spelling issues. In [10] proposed an English spelling correction model for Arabic users students who need English in their academic writing based on n-gram technique. The students who speak Arabic has difficulty in using English in writing reports and other academic document in their study. This model aims to help those students to write English better by having less word spelling mistakes. In [2] studied on different approaches of spelling error detection and correction. They used n-gram with dictionary lookup approaches for detection. For the correction they use the approaches of neural based, edit distance, similarity keys, rule-based, n-gram, and probabilistic. Another research [11] explored the possibility of using neural models for spelling correction in Azerbaijani, an agglutinative language with complex word structures and rich morphological features. In [12] used the Expectation Maximization (EM) algorithm to learn edit distance wight directly from the search query logs. The error model, which is a weighted string edit distance measure, is typically learned from pairs of misspelled words and their corrections. Instead of relying on corpus of paired words, the paper proposed learning the edit distance wights from search query logs using the EM algorithm. This approach allows for the learning of an accurate error model without the need for a corpus of paired strings. The [13] observed on common misspelling correction of users when they search for a product online. While other works rely on either human annotated or the entire web, this work used users' event logs from an e-commerce website to fetch similar search query pairs within an active session. The main idea is that the users issue a search query but alter it into something similar within a given time window which might be the correction of a potential typo. In [14], Google introduced a language model BERT, which stands for Bidirectional Encoder Representations from Transformers. Unlike the traditional approach that trains the models word by word, BERT can train the models based on the complete set of words in a sentence. In [15] studied on encoding processes involved in discriminating and recognizing homophones, synonyms, and unrelated words, and how phonetic and semantic information is retrained and utilized in these tasks in English. As the result, unrelated words were recognized better than homophones or synonyms, and paired presentation led to lower discrimination of unrelated words compared to single presentation. The research [16] aims to address the detection and correction of real-word errors in Bangla text, focusing on homophone errors, using a combination of bi-gram and tri-gram models. As a result, they achieved 96% accuracy in detection and correction real-word errors of Bangla text with three groups of corpora that contain: a collection of sets of homophones (confusing) words, the collection of bigrams and trigrams using homophone words, and the test set.

Beside the above studies, there are also a few research related to homophones and spelling correction in Khmer. In [17] studied on detection and correction of homophonous non-word error for Khmer language using Khmer Common Express (KCE). The result evaluation was 92.43% accuracy from comparing the number of errors correctly detected to total number of misspellings. In [18] proposed a Khmer word segmentation using a new algorithm with ternary decomposition technique to extract new combination words that are used in daily life while the current Choun Nath's Dictionary[1] that most research in Khmer relies on keeps keywords only. With a ternary decomposition technique, this work achieved 88.8%, 91.8%, and 90.6% rates of precision, recall and F-measurement. Another work [19] proposed a word spelling correction model for ancient Khmer language within the 18th century. It roles as a spell checker and spelling correction to speed up the digitization process of [20]. In [3] proposed a model called Khmer Spelling Checker (KSC) that integrated with Microsoft Word to help students in Travinh University who have difficulty in making report on Khmer language subject. The result shows that KSC achieved 97% of relatively high results in comparison with related previous research.

In summary, the literature examined demonstrates various methods for dealing with these challenges, which include n-gram techniques, neural models, and advanced algorithms such as the EM algorithm and BERT. These techniques have proven to be effective in multiple languages, such as English, Azerbaijani, Bangla, and Khmer, illustrating the universal significance of correcting spelling. Additionally, studies on homophones and real-word errors highlight the complexity of the issue, especially in languages with intricate morphological characteristics. Initiatives to create spelling correction tools, like the Khmer Spelling Checker (KSC), underscore the potential of these technologies to improve communication and educational outcomes. In conclusion, continuous research and progress in this field are essential for enhancing the precision and impact of written communication in diverse languages and contexts.

---

[1] Choun Nath's Dictionary is the ultimate reference for Cambodian language or Khmer. It was first published in 1938 and become an important document that cannot be overlooked for Khmer.

## 4 Methodology

To define the significance of homophones spelling correction model and propose the most appropriate solution, we come up with two important works. Firstly, we conducted a survey with questionnaires that were specifically designed for identifying the challenges and the need of Khmer authors in utilization of homophones in Khmer writing. Then we look at the current related works to define the most appropriate model that can be used to fix our current problem.

### 4.1 The Survey

We carried out a survey aimed at Khmer native speakers who use the language in their everyday lives. The survey includes seventeen questions, detailed in the appendix below, organized into five sections as follows.

The first section we started from background of participants where we gathered information of participants concerning on their educational levels and experiences with the Khmer language, providing insight into their expertise. Understanding their backgrounds ensures survey accuracy. Then we continued to explore their experiences with Khmer homophones in section 2. We examined their familiarities, common challenges, their confidence in identifying homophones, and their strategies for managing homophone errors when they faced those issues. In section 3, we addressed on how homophone errors affect the clarity and communication of participants' Khmer writing. We assessed whether homophones are a significant challenge or can be overlooked. After that, we surveyed on the current existing tools related to homophone concern in Khmer. This section focused on the current tools that participants used to address homophone errors, providing insights into the tools and methods available and the features participants would find beneficial in future solutions. Finally, we gathered open feedback from participants in section 5. This last section allowed participants to share their additional feedback including specific challenges, needs, and suggestions that could inform the development of a more effective homophone correction model.

In conclusion, this study allowed us to examine participants' backgrounds and experiences with Khmer homophones to better understand the writing challenges they encounter. It helped to identify common difficulties, evaluate the impact of homophone errors on communication, review current tools, and gather feedback on the need for enhanced correction tools. The results highlight the importance of developing a targeted homophone correction model to improve clarity in Khmer writing.

### 4.2 The Current Works

The current existing studies role as a significant part to help us define and propose an appropriate solution to help authors related to their homophones concern. We go through the previous works which start from the studies in spelling correction in rich resource language like English, French, and Arabic to low resources language like Khmer. Then we go through more specifically to our research which is related to homophones concern in different languages, a lastly to Khmer language. Additionally, the language model is another important component that deserves our effort to go through which can be proposed a new solution for the current problems. With these activities it helps us to understand the current works, solutions, and reflect on our works and be able to propose an appropriate solution for the problem.

## 5 Result and Discussion

### 5.1 Background of Participants

There are 108 Khmer native speakers involved in our survey to understand the challenges of homophones in their Khmer writing and help us to propose an appropriate solution to the current need. There are three groups of participants in our survey including 57.41% are students which is the primary group, then 39.81% are employees, and 2.78% are businessmen (Figure 1.a). For their educational level, 6.48% finished high school only, 65.74% are studying or studied at the university, and 27.78% pursued higher degrees including master's and PhD degrees

(Figure 1.b). The major group of participants are from the field of science and technology which equal to 69.44%, followed by 30.56% from the field of social science which included Khmer literature, business and communications, accounting, finance, and banking (Figure 1.c).

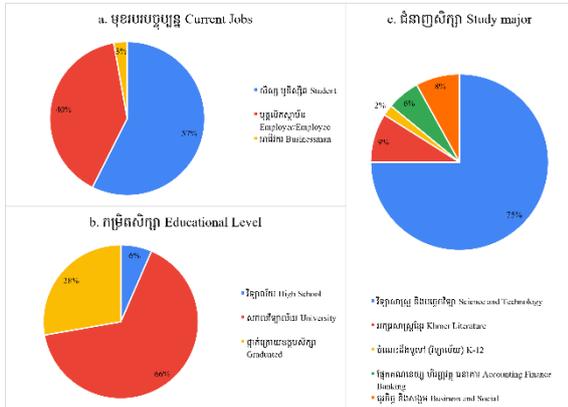

Figure 1. Background of participants

To make our study more precise, we also observed their frequency of using Khmer language in daily life by scoring themselves from 0 represent to never write Khmer to 5 represent to always writing Khmer. The result showed that 34.41% rated themselves as 5 for always write Khmer, 24.07% rated themselves as 4, 26.85% rated themselves as 3, 12.96% rated themselves as 2, 4.63% rated themselves as 1, and no one rated themselves as 0, never write Khmer. This data assures our survey accuracy that our survey is conducted to the right group of audience which require participants that experiencing writing Khmer. In addition, we also observed the types of articles that they primarily write Khmer in their daily life. This is a multiple answers question which allows participants to select more than one answer that applies to their situation. The result shows that the top three Khmer writing that apply to participants situation are writing on social media 45.37%, academic papers 41.47%, and the same rate of writing to activity reports. The rest three categories that our participants also mentioned are technical documents 22.22%, communication in chat the same as other notes 2.78%.

### 5.2 Experience with homophones in Khmer of participants

In section 2, we defined the participants' experience with homophones in Khmer by conducting several questions. The first question, we surveyed on how often they encounter homophones in their Khmer writing. By voting from 0 represent to never encounter homophones, to 5 represent to always encounter homophones, we got the result as follows. 16.67% voted for 5, 22.22% voted for 4, 38.89% voted for 3 which is the highest voted among others, 17.59% voted for 2, and 4.63% voted for 1, and 0.93% voted for 0, never encounter homophone in Khmer. With a similar method, the next question we defined how confident participants to in are identifying the correct Khmer homophones based on their context. The result showed that there are 8.33% voted for fully or always confidence to identify the homophones based on the context where it is in, 33.33% voted for 4, 42.59% voted for 3 which demonstrates that they are in neutral or not very confidence when identify the homophone where it is located, and 12.04%, 4.63%, 0% voted for 2, 1, and 0 respectively (Figure 2).

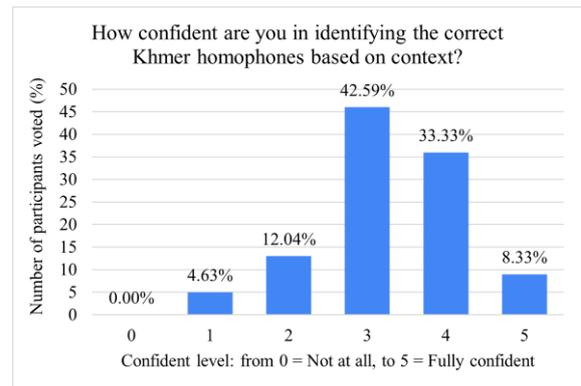

Figure 2. Survey results of authors' confidence in using correct homophones in Khmer writing

We conducted another two questions in the same section by allowing participants to be able to select more than one answer that applies to their situation. We inspect the most confusing homophones in Khmer and how they deal with those problems. The result indicated that the most confusing homophones are 54.63% with the terms "អនុវត្ត" and "អនុវត្តន៍", 47.22% with the terms "ក" and "ការ" and "ការណ៍", 23.41% with the terms "ផ្សា" and "ផ្សារ". Another result showed how authors deal with the homophones problems with 60.19% choose to rely on their knowledge, 28.70% choose to ask someone else to check, 53.70% choose to use spelling and grammar checker tool, and another 5.56% ignore the problem by not identifying them.

## 5.3 Impacts of Homophone Spelling Errors to Authors in Khmer Writing

In this section, we focused on the challenges of homophone spelling errors to authors in their Khmer writing. The first question – have homophone errors ever led you to miscommunication or misunderstanding with your Khmer readers? There are four possible answers: yes often, yes sometimes, not sure, and never. The result showed that 69.44% responded to yes sometimes, followed by 16.67% said that homophone errors often led to miscommunication, 7.41% were not sure about the answer, and 6.48% responded to never. The next question we studied on how often to homophones errors impact the clarity of their Khmer writing, by asking participants to score themselves from 0 represents to never, to 5 represents to always. The result showed that 7.41% scored themselves to 5 refer to always, 22.22% scored to 4, 30.56% scored to 3, 22.22% scored to 2, 15.74% scored to 1, and lastly 2.78% scored to 0 which say that homophones errors never impact to the clarity of their works. The last question in this section we studied on how homophones spelling errors in Khmer affect their writing process by allowing participants to be able to select more than one answer. The result showed that 77.78% claimed that it makes them doubt word selection or make confusion to them, 33.33% claimed that it slows down their writing process, and 30.56% claimed that it disrupts their writing flow.

## 5.4 Current Existing Tools for Homophone Errors Correction in Khmer

In this section we observed the current existing tools on Khmer homophone errors correction as well as how authors deal with those problems when they encountered them. The first question was – Are you using any tools or techniques to assist with homophone mistakes in your Khmer writing? The results indicated that 68.52% of respondents confirmed they neither use nor have any methods to address homophone errors in Khmer writing. The rest of 31.48% do not use any homophone spelling errors correction tools, but they use Khmer dictionary, Gboard (Google Keyboard), and Google translate to deal with Khmer homophone errors in their writing (Figure 3).

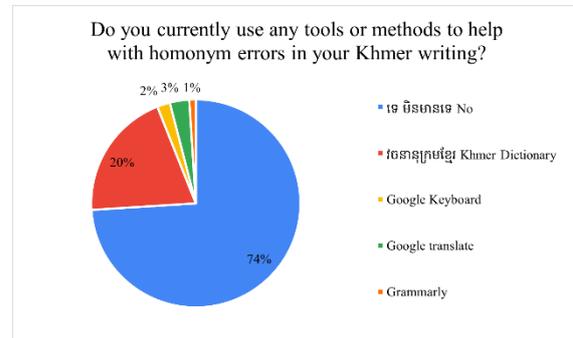

Figure 3. Responds of participants to the current existing tool for homophone spelling correction in Khmer

Furthermore, we also observed how effective those methods are if they have used any. Among those who responded they use any above methods, 12.04% confirmed that it is not helpful at all which rated to 0, 1.85% rated for 1, 9.26% rated for 2, 25.00% rated for 3, 21.30% rated for 4, and only 11.11% rated for 5 which is always useful to them. The last question in section 4, we went a bit further to if there is a tool for homophone spelling correction for Khmer, what features do authors want to see in that model by allowing them to give more than one answer to the multiple-choice selection. Most of the participants 73.15% want to see tool for homophone automatic correction for Khmer language, 47.22% want to see contextual analysis for correct homophone usage, 21.30% want to have an integration with their current writing software like Microsoft Words, and Google docs, 27.78% want to have a feature of feedback and learning resources about common Khmer homophone errors, and 23.15% want to see a feature of customizability of suggestions.

## 5.5 Open feedback from participants

The last section of this survey, we proposed three questions to help participants to give feedback related to their problems in utilization of homophones in Khmer writing. The first question observed on other challenges that Khmer authors face in their writing. We received similar answers indicating that homophones in Khmer are quite confusing. They have remarkably similar spelling, the same pronunciation, but different meanings which causes difficulties in choosing the right words in the right context in their writing. In addition, the most challenge homophones that most authors mentioned are the terms with consonants "ʳ" (example: "ផ្សា" and "ផ្សារ") and "ិ" (example: "អនុវត្ត" and "អនុវត្តន៍"). However, they have not found any applications or related studies to help them overcome these problems yet. The next question, we asked participants to imagine if there is a homophone spelling checker application for Khmer, how do they think that tool could best support in their writing process? All participants agree that such a tool would help them use homophones more accurately, boost their confidence in writing, and enhance their overall Khmer writing process. Additionally, they believe it would accelerate their writing tasks and contribute to the promotion of the national language. In the final question of our survey, participants were invited to share any additional comments or suggestions about homophone challenges in Khmer writing. Most respondents expressed a desire for an application specifically designed to handle homophone usage in Khmer, while others suggested that such a tool should be compatible with existing applications like Microsoft Word and Google Docs. Finally, participants also expressed a desire for an explanation feature in the homophone correction tool, which would help users understand and improve their knowledge of the correct usage of those terms.

### 5.6 Result of the Related Studies

The literature review in Section 3 reveals a range of studies on spelling correction and homophones across various languages, though most focus primarily on correcting general misspellings. There is different research related to spelling correction and homophones. However, most studies focus on word misspelling correction alone. For example, in [11] and [13] investigated on word suggestion and spelling correction in Azerbaijani and English respectively. Other works focused on the same problems, except in different languages. Another interesting study [14], Google introduced a language model called BERT that can perform misspelling correction based on the sentence in English. We noticed that there were other works studied related to homophones in English and Bangla such as [15] and [16] respectively. In Khmer, there are few research related to spelling correction such as [3] and [18], and [17] on Khmer homophone correction.

While many studies have focused on spelling correction, there is a noticeable lack of research addressing homophone-related issues, particularly in the Khmer language. Homophones are often confused, leading to unintentional writing errors, and they present a greater challenge than typical typographical mistakes.

### 5.7 Discussion

The survey results demonstrated that homophones are a common source of confusion for Khmer writers. Despite their varying educational backgrounds and experiences, most participants reported encountering homophones frequently and struggling to identify them correctly based on context. This issue is further compounded by the lack of tools specifically designed to assist with homophone errors in the Khmer language. While some participants reported using general tools like Khmer dictionaries and Google Translate, they are evident that more resources are needed to address the specific challenges in homophone pose. In our literature review, we found a few pieces of research related to the concern of homophone spelling correction. However, there has yet to be a specific study on this problem in Khmer. This gap in research underscores the need for more tools specifically designed to assist with homophone errors in the Khmer language, and it is an area that deserves our collective effort.

## 6 Conclusions

Khmer homophones pose a substantial challenge to the clarity of written communication, as authors often struggle to distinguish between similar-sounding words, leading to misunderstandings and diminished confidence in their writing. Both survey responses and a literature review revealed that Khmer writers frequently encounter homophones and have difficulty choosing the correct word due to the lack of effective tools. Current resources, such as Khmer dictionaries and Google Translate, do not adequately address these specific challenges. Furthermore, the limited research focused on Khmer homophones underscores the urgent need for a specialized tool with contextual analysis to improve accuracy. Addressing this gap is crucial for creating effective resources to help writers manage homophone errors in Khmer.

In conclusion, this study underscores the importance of developing specialized tools to assist with homophone errors in the Khmer language. By bridging the current gap in research and resources, such a tool would empower Khmer authors to write with greater confidence and precision, contributing to the enrichment and preservation of the language. Continuing effort in this area is essential to ensure that Khmer, like other languages, can fully benefit from advancements in technology and linguistics. The proposed tool has the potential to significantly improve the quality of written communication in the Khmer language, thereby enhancing the overall language experience for its users.

## Appendix: The Survey Questions

**Section 1: Background of Participants**

1. What is your current job?
2. What is your educational level?
3. What is/was your study major?
4. How often do you write Khmer?
5. What type of content do you primarily write in Khmer?

**Section 2: Participants' Experiences with Homophones in Khmer**

6. How frequently do you encounter homophones in your Khmer writing?
7. How confident are you in identifying the correct Khmer homophones based on context?
8. Which Khmer homophones do you find most challenging to differentiate?
9. How do you usually identify and correct homophone errors in your writing?

**Section 3: Impacts of Homophones Errors in Participants' Writings**

10. How often do homophone errors impact the clarity of your Khmer writing?
11. How do homophone errors in Khmer affect your writing process?

**Section 4: The Current Existing Tools Related to Homophone Concern in Khmer**

12. Are you using any tools or techniques to assist with homophone mistakes in your Khmer writing? If yes, please specify.
13. If you have used them, how effective are these tools or methods helping you with Khmer homophone errors?
14. What features would you find the most helpful in a tool designed to assist with Khmer homophone errors?

**Section 5: Open Feedback from Participants**

15. What are the biggest challenges you face with homophones in Khmer writing?
16. How do you think a homophone correction tool could best support your Khmer writing process?
17. Any additional comments or suggestions regarding homophone challenges in Khmer writing and potential solutions?